\documentclass[times, review, 10pt]{elsarticle}

\usepackage{amssymb}
\usepackage{amsmath}

\journal{Pattern Recognition}

\usepackage{placeins}
\usepackage{url}
\usepackage{multirow}
\usepackage{booktabs}
\usepackage{adjustbox}
\usepackage{caption}
\usepackage[table,xcdraw]{xcolor}
\usepackage{tikz}
\usepackage{gensymb}
\usepackage{tabularx}
\newcolumntype{C}[1]{>{\centering\arraybackslash}p{#1}}
\usepackage{orcidlink}
\usepackage{amsmath}
\usepackage{pifont}
\usepackage{subcaption}
\usepackage{placeins} 
\usepackage{float}

\begin{document}

\begin{frontmatter}



\title{Learning Generalizable and Efficient Image Watermarking via Hierarchical Two-Stage Optimization}


\affiliation[label1]{organization={Shenzhen Institute for Advanced Study, University of Electronic Science and Technology of China},
                     city={Shenzhen},
                     country={China}}
\affiliation[label3]{organization={School of Computer Science and Technology, Tongji University},
                     city={Shanghai},
                     country={China}}
                     
\affiliation[label2]{organization={Bytedance},
                     city={Hangzhou},
                     country={China}}

\author[label1]{Ke Liu} 
\author[label1]{Xuanhan Wang}
\author[label2]{Qilong Zhang}
\author[label1]{Lianli Gao}
\author[label3]{Jingkuan Song}

\begin{abstract}
Deep image watermarking, which refers to enabling imperceptible watermark embedding and reliable extraction in cover images, has been shown to be effective for copyright protection of image assets.
However, existing methods face limitations in simultaneously satisfying three essential criteria for generalizable watermarking: 1) \textbf{invisibility} (imperceptible hiding of watermarks), 2) \textbf{robustness} (reliable watermark recovery under diverse conditions), and 3) \textbf{broad applicability} (low latency in the watermarking process). To address these limitations, we propose a \textbf{H}ierarchical \textbf{W}atermark \textbf{L}earning (\textbf{HiWL}) framework, a two-stage optimization that enables a watermarking model to simultaneously achieve all three criteria.
In the first stage, \textbf{distribution alignment learning} is designed to establish a common latent space with two constraints: 1) visual consistency between watermarked and non-watermarked images, and 2) information invariance across watermark latent representations. In this way, multi-modal inputs—including watermark messages (binary codes) and cover images (RGB pixels)—can be effectively represented, ensuring both the invisibility of watermarks and robustness in the watermarking process.
In the second stage, we employ \textbf{generalized watermark representation learning} to separate a unique representation of the watermark from the marked image in RGB space. Once trained, the HiWL model effectively learns generalizable watermark representations while maintaining broad applicability.
Extensive experiments demonstrate the effectiveness of the proposed method. Specifically, it achieves \textbf{7.6\% higher accuracy} in watermark extraction compared to existing methods, while maintaining extremely low latency (processing 1K images in \textbf{1 s}). Our code is publicly available at \textit{\url{https://github.com/xxykkk/HiWL}}.
\end{abstract}




\end{frontmatter}


\section{Introduction}

In recent years, we have witnessed tremendous success in artificial intelligence generated content (AIGC). Numerous high-quality visual assets such as images or videos can be created effortlessly. Dramatic increase of visual assets, however, raises a wide public concern about violations to copyright ownership of these assets. Deep image watermarking, a deep learning based approach to hide a piece of ownership message into an image, has become a key solution to provide protection of copyright ownership. 

\begin{figure}[t]
    \centering
    \includegraphics[width=0.65\linewidth]{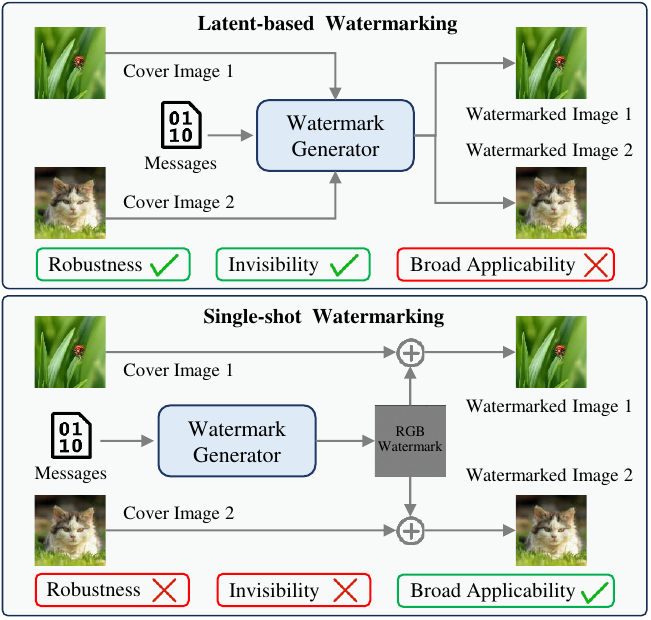}
    \caption{Deep watermarking paradigms including the latent-based and the single-shot. Existing methods do not satisfy invisibility, robustness, and broad applicability simultaneously.}
    \label{fig:1}
\end{figure}

As illustrated in Fig.~\ref{fig:1}, existing methods primarily follow encoder-decoder frameworks, manifesting as two paradigms: latent-based methods \cite{hidden, two-stage, mbrs, flow-based, dwsf} and single-shot methods \cite{udh}. Specifically, the latent-based methods often start with latent representation fusion between watermark message and the cover image by the encoder, and then perform watermarked image generation and message extraction, respectively, via the decoder and extractor. In a sharp contrast, the single-shot series \cite{udh} believe the watermark fusion should happen in RGB pixel space rather than latent space. They often initially transform the original message to image-like watermark via an encoder-decoder module and then perform a simple summation over the cover image and transformed messages for watermark image generation. 

\begin{figure}[tbph]
    \centering
    \includegraphics[width=0.55\linewidth]{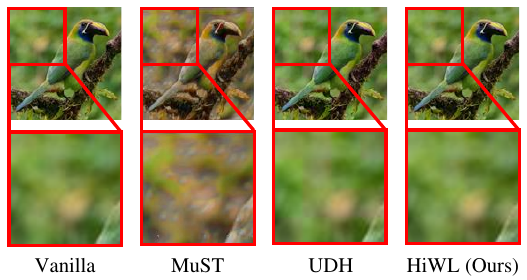}
    \caption{Visualization of generated watermark images, which are separately produced by UDH \cite{udh}, MuST \cite{must} and Ours. \textit{vanilla} indicates the cover image.}
\captionsetup{justification=centerlast}
    \label{fig:2}
\end{figure}

Despite the success achieved to date, we argue that a good image watermarking model in real-world applications should satisfy three criteria: \textbf{invisibility}, \textbf{robustness}, and \textbf{broad applicability}. The invisibility requires hiding watermark messages while keeping the watermark images look natural. The robustness ensures precise watermark extraction in serious distortion situations. The broad applicability refers to the low latency of the watermarking process, making it suitable for large-scale images or videos. However, existing methods are not capable of satisfying all three criteria. 
In terms of robustness and invisibility, latent-based methods typically perform better thanks to their representation fusion \cite{mbrs,dwsf}, whereas single-shot methods frequently exhibit poor robustness and produce visible watermark blocks, as illustrated in Fig.~\ref{fig:2}.
In terms of broad applicability, single-shot methods often perform better than the latent-based ones, as the latter suffer from high latency. This is because watermarking in a latent-based paradigm requires an encoder module to perform information fusion for each new cover image, making it unsuitable for large-scale image processing.
However, in single-shot pipeline, the watermarks are only hidden in the RGB space, making it difficult to recover the messages from the watermark image when encountering realistic distortions, which leads to poor message recognition performance. 
Hence, this raises a critical question: \textbf{\textit{is it possible to design a watermark model satisfying all three criteria, and how can it be achieved?}}

In this paper, we seek to answer the aforementioned question by proposing \textbf{H}ierarchical \textbf{W}atermark \textbf{L}earning (\textbf{HiWL}), the first single-shot paradigm that can satisfy three criteria simultaneously. Specifically, the HiWL involves a novel two-stage optimization of watermark generation: 1) distribution alignment, and 2) generalized watermark representation learning. 
In the first stage, we perform multi-level alignment between the cover images and messages through latent representation fusion, ensuring both \textbf{robustness} and \textbf{invisibility}.
In the second stage, we propose multi-image adaptation and asynchronous optimization for generalized representation learning, which penalizes large variations of the watermarks across different cover images. Specifically, we use the RGB residual between a reference cover image and a watermark image pair as a generalized watermark representation. After that, this residual is added to other images to generate numerous watermarked images. During training, those marked images are utilized to learn generalizable message pattern for \textbf{broad applicability}. 
In summary, the main contributions are three folds:
\begin{itemize}
    \item We systematically reveal the shortcomings behind existing watermarking paradigms (i.e., latent-based and single-shot), which fail to simultaneously achieve robustness, invisibility, and broad applicability.
    \item We propose \textbf{H}ierarchical \textbf{W}atermark \textbf{L}earning (\textbf{HiWL}), a generalizable single-shot watermarking framework that employs distribution alignment to ensure robustness and invisibility, and utilizes generalized watermark representation learning to achieve broad applicability.
    \item We perform extensive experiments to demonstrate that the proposed HiWL achieves superior results compared to existing state-of-the-art approaches across 18 distortion settings. Furthermore, it attains exceptionally low latency, processing 1,000 images in just 1 second for watermarking.
\end{itemize}

\section{Related Work}
\subsection{Handcrafted Image Watermarking}

Handcrafted image watermarking has evolved significantly from early spatial-domain methods \cite{lsb,lsb2} to frequency-domain techniques using DCT, DFT, and DWT \cite{dct1,dwt2}. Building on these foundations, Tsai et al. \cite{pr-5} propose to balance transparency and robustness by integrating Just Noticeable Difference (JND) profiles in the wavelet domain. Furthermore, Shen et al. \cite{pr-4} introduce an associative watermarking approach, which combines data mining rules with similarity diagrams to ensure watermark uniqueness. To enhance watermark detection, Etemad et al. \cite{pr-3} employ the $t$ Location-Scale distribution for modeling contourlet coefficients, deriving a multiplicative detector via the Likelihood Ratio Test. Together, these methods demonstrate a clear shift toward perceptually aware and statistically sophisticated watermarking frameworks. Compared to current deep watermarking methods, they exhibit significantly poorer robustness.

\begin{figure*}[t]
	\centering
	\includegraphics[width=0.95\linewidth]{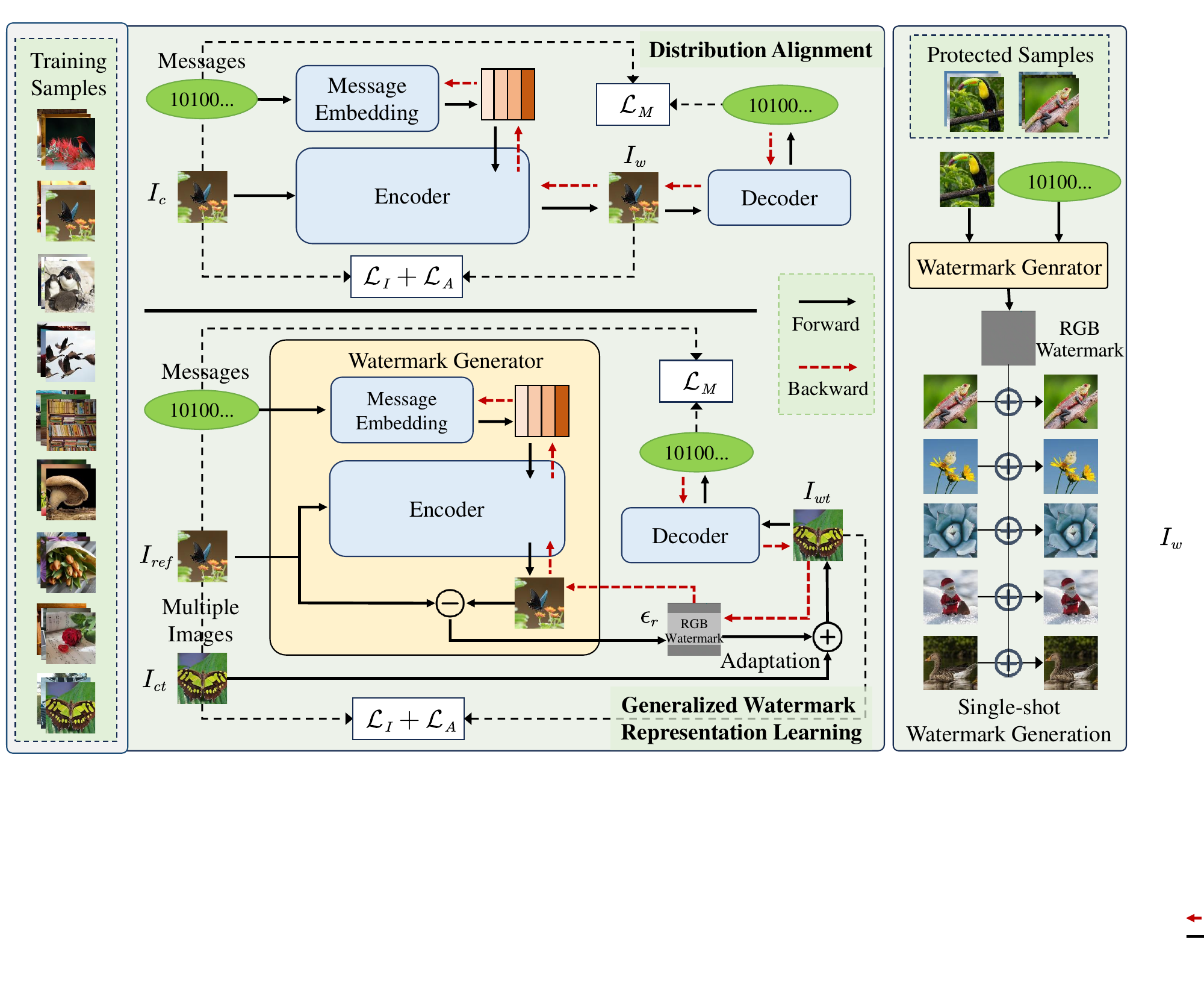}
	\caption{The overview of proposed HiWL in the training phase (left) and inference phase (right). During training, it involves a two-stage optimization. In the first stage, the image reconstruction loss $\mathcal{L}_I$, adversarial loss $\mathcal{L}_A$, and message reconstruction loss  $\mathcal{L}_M$ are jointly used for multiple alignment between cover image and watermark messages. In the second stage, multi-image adaptation is designed to learn generalized RGB watermarks.}
	\label{fig:framework}
\end{figure*}

\subsection{Deep Learning based Image Watermarking}
The remarkable success of deep learning has significantly advanced a wide range of visual recognition tasks \cite{friends-1, friends-2, friends-3}, including digital watermarking. 
Early learning-based watermarking methods \cite{hidden, flow-based, tmm-7, dwsf, resolution, sslw} primarily rely on latent-based frameworks. To improve watermarking efficiency, Zhang et al. \cite{udh} propose a single-shot Universal deep hiding method. Both approaches are based on encoder–decoder frameworks. 
To further improve robustness under realistic distortions, subsequent works~\cite{two-stage,mbrs} explore training strategies tailored to specific noise models. 
Inspired by the success of artificial intelligence generated content (AIGC), a growing number of studies have begun to frame image watermarking as an image generation task, giving rise to two predominant paradigms: GAN-based~\cite{gan1, gan2, gan3} and diffusion-based approaches~\cite{ldm, ldm2, ldmx, robin, meng2024latent}.
Despite their success, existing image watermarking methods still fall short of simultaneously meeting all three criteria.

\section{Methodology}

\subsection{Problem Formulation}
The goal of watermarking is to recover copyright messages from a crafted image with a visually imperceptible watermark. Conventionally, the watermark model is built upon an encoder-decoder architecture. 
Formally, let $I_{co} \in \mathbb{R}^{3 \times h \times w}$ denote the cover image
with spatial resolution $h \times w$, and let $M \in \{0,1\}^{L}$ denote a
watermark message of length $L$. 
The watermarking aims to craft an image ${I}_{wm}$ that has messages and satisfies following constraint:
\begin{equation}
\begin{aligned}
\mathcal{F}_{en}(I_{co}, M) & = I_{wm}, \\
\textit{s.t.} \quad
 \|{I}_{wm} & -{I}_{co}\|_{\infty}  \leq\epsilon,
\end{aligned}
\label{eq:watermarking}
\end{equation}
where $\mathcal{F}_{en}(\cdot)$ denotes the watermark encoder that injects imperceptible messages into the image. $I_{wm}$ represents the watermark image generated by the encoder. The $\epsilon$ is the perturbation hyper-parameter that constraints similarity between the cover image and the watermarked image. Furthermore, the hidden message should be reliably recovered via the decoder, as illustrated in Equ.~\ref{eq:decoder}:
\begin{equation}
\begin{aligned}
\mathcal{F}_{de}(I) & = M_{ex},  \\
\textit{s.t.} \quad
 M_{ex} & \approx M,
\end{aligned}
\label{eq:decoder}
\end{equation}
where $\mathcal{F}_{de}(\cdot)$ is the decoder, and $M_{ex}$ represents the recovered message via the decoder. 

\begin{figure}
    \centering
    \includegraphics[width=0.7\linewidth]{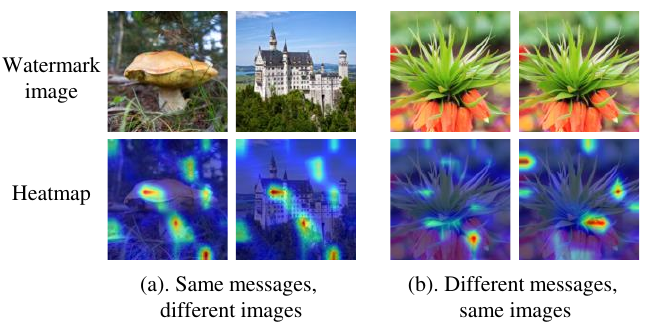}
    \caption{Visualization of gradient heatmaps in the decoder $\mathcal{F}_{de}(\cdot)$.}
    \label{fig:motivation}
\end{figure}

\subsection{Hierarchical Watermark Learning}

In the general training process, two objectives are typically adopted to optimize the watermark models: 1) message reconstruction, and 2) image reconstruction. The former aims to ensure the completeness of the hidden watermark messages, while the latter is designed to ensure the invisibility of the messages. 
As shown in Fig.~\ref{fig:motivation}, we observe a weak association between the image and its watermark in the latent space. Theoretically, this can be attributed to the fact that constraints 1 and 2 do not explicitly couple the watermark message with the image features. If this link can be completely severed, a general-purpose watermark representation, which is applicable to any image, could be obtained. Motivated by this observation, we propose a hierarchical two-stage optimization strategy, including distribution alignment and generalized watermark representation learning, as shown in Fig.~\ref{fig:framework}.

\subsubsection{Distribution Alignment}
\begin{figure*}[h]
	\centering
	\includegraphics[width=0.98\linewidth]{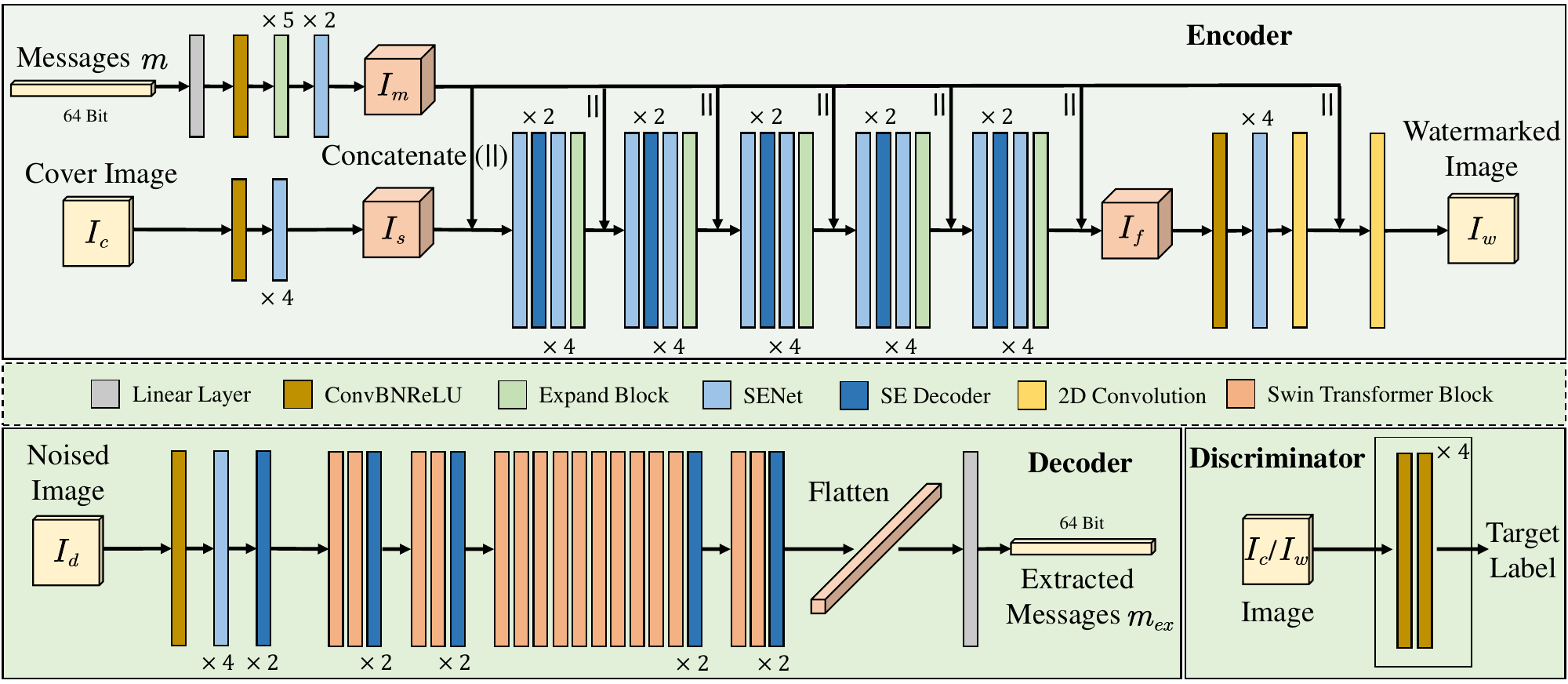}
	\caption{The structure diagram of the proposed method (HiWL), which consists of an encoder, a decoder, and a discriminator.}
	\label{fig:model}
\end{figure*}

In the first phase, we follow the general training process. As illustrated in~Fig.~\ref{fig:model}, both the message and cover image are fed into the encoder for joint training.
Formally, we denote the cover image as $I_c$, the messages as $M$, and the watermarked image as $I_w$. The image reconstruction learning objective $\mathcal{L}_I$ is defined through Equ.~\ref{eq:encoderloss}:
\begin{equation}
    \mathcal{L}_{I} = MSE(I_c,I_w) + \mathcal{S}(I_c,I_w),
    \label{eq:encoderloss}
\end{equation}
where $\mathcal{S(\cdot)}$ is standard \textit{ssim\_loss} function proposed in \cite{mssim}. Furthermore, we adopt mean square error as the message reconstruction loss function $\mathcal{L}_M$:
\begin{equation}
    \mathcal{L}_{M} = MSE(M, \mathcal{F}_{de}(I_w)).
    \label{eq:decoderloss}
\end{equation}
Beyond the standard loss functions used to optimize the watermark model, we adopt an adversarial loss term, $\mathcal{L}_A$, to augment the visual quality of the generated images:
\begin{equation}
	\begin{array}{lll}
		\mathcal{L}_A = log(\mathcal{F}_{a}(I_{c})) + log(1-\mathcal{F}_{a}(I_w)),&
	\end{array}
	\label{eq:advloss}
\end{equation}
where $\mathcal{F}_a$ denotes the discriminator function.
Accordingly, the overall learning objective in the first stage can be formulated as Equ.~\ref{eq:totalloss}:
\begin{equation}
	\begin{array}{lll}
		\mathcal{L} &=\mathcal{L}_{I}+ \mathcal{L}_{M}+ \mathcal{L}_A.&\\
	\end{array}
	\label{eq:totalloss}
\end{equation}
The whole process is trained using end-to-end optimization based on the loss function $\mathcal{L}$.

\subsubsection{Generalized Watermark Representation Learning}
In the second stage, we first select a cover image as the reference, which is denoted as $I_{ref}$. Given watermark message $M$, we then compute the residual between the reference image and its watermarked counterpart:
\begin{equation}
	\begin{array}{lll}
		\epsilon_{r} &= I_{ref} - \mathcal{F}_{en}(I_{ref}, M), &\\
	\end{array}
	\label{eq:residual}
\end{equation}
where $\epsilon_{r}$ denotes the residual, and we regard it as the watermark message representation. For watermark embedding, the residual can be directly applied to an image $I_{ct}$, yielding its watermarked counterpart $I_{wt}$:
\begin{equation}
	\begin{array}{lll}
		I_{wt} &= I_{ct} + \epsilon_{r} &\\
	\end{array}
	\label{eq:residual_mark}
\end{equation}
To enable this residual representation generalize to any input image, we apply three aforementioned loss functions to the generated watermarked image, as formulated in Equ.~\ref{eq:totalloss2}:
\begin{equation}
	\begin{array}{lll}
		\mathcal{L}^{’} &=MSE(I_{ct},I_{wt}) + \mathcal{S}(I_{ct},I_{wt})+ MSE(M, \mathcal{F}_{de}(I_{wt})).&\\
		                & + \quad log(\mathcal{F}_{a}(I_{ct})) + log(1-\mathcal{F}_{a}(I_{wt})) &
	\end{array}
	\label{eq:totalloss2}
\end{equation}

After training, the deep watermark model is invoked only once to generate the residual representation for each watermark message. For practical use, the watermark is embedded by simply adding this pre-computed residual to a target image, offering a one-shot solution with broad applicability.

\section{Experiments}


\subsection{Experimental Settings}
\textbf{Datasets.}
We utilize the MS-COCO dataset \cite{coco} for training. For a comprehensive evaluation, ImageNet \cite{imagenet}, HumanArt \cite{humanart}, DRCT \cite{drct}, and GenImage \cite{genimage} are jointly leveraged to complement the test dataset. In particular, HumanArt is a multi-style image covering movies, anime, reality, cartoons, etc. DRCT and GenImage are image datasets generated by various generative models. During the training phase, we use 50{,}000 images and 100{,}000 images as Stage 1 and Stage 2 training sets, respectively. 

\begin{table}[t]
	\centering
	\caption{Noise factors on training and testing.}
	\label{tab:para_noise}
	\resizebox{0.8\linewidth}{!}{%
		\begin{tabular}{lll|lll@{}}
			\toprule
			Noise & Train & Test & Noise & Train & Test \\ \midrule
			JPEG & $q\in[40,100]$ & $q=50$ & Contrast & $p\in[-0.8,0.8]$ & $p=\pm 0.8$ \\
			GN & $\sigma \in[3,10]$ & $\sigma=10$ & Resize & $p\in[-0.5,0.5]$ & $p=\pm0.5$ \\
			GF & $\sigma \in[3,8]$ & $\sigma=8$ & Crop & $p\in[0.7,1]$ & $p=0.7$ \\
			Dropout & $p\in[0.7,1]$ & $p=0.7$ & PIP & $p\in[0.25,1]$ & $p=0.25$ \\
			MF & $\sigma \in[0,7]$ & $\sigma=11$ & Padding & $p\in[0,50]$ & $p=50$ \\
			Color & $p\in[-0.5,0.5]$ & $p=\pm0.9$ & Occlusion & $p\in[0.0625,0.25]$ & $p=0.25$ \\
			Bright & $p\in[0,0.5]$ & $p=0.5$ & Rotate & $r\in[0,360]$ & $r=180$ \\
			Saturation & $p\in[-0.8,0.8]$ & $p=\pm0.9$ & Shear & $s\in[0,30]$ & $s=30$ \\
			Hue & $p\in[-0.7,0.7]$ & $p=\pm0.6$ & Affine & $r\in[0,360],s\in[0,30]$ & $r, s=180, 30$ \\ \bottomrule
		\end{tabular}%
	}
\end{table}
\begin{figure}[t]
	\centering
	\includegraphics[width=0.85\linewidth]{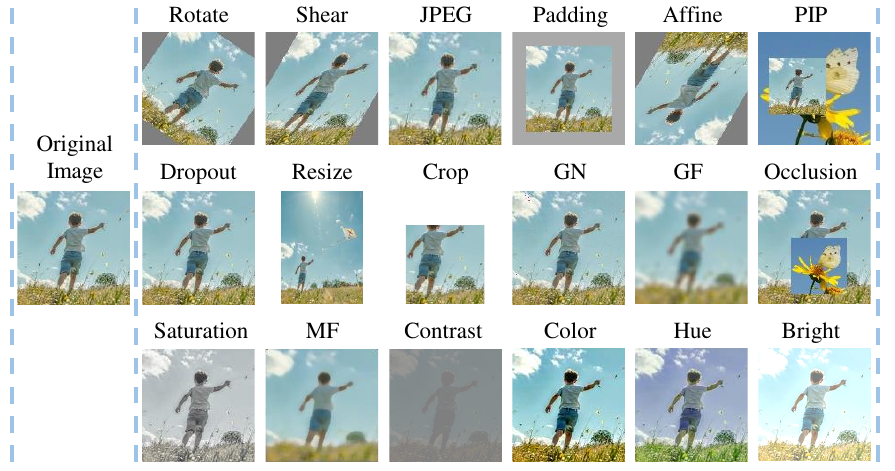}
	\caption{Visualization of 18 types of noise distortions with random strength factors.}
	\captionsetup{justification=centerlast}
	\label{fig:noise}
\end{figure}

\begin{table*}[t]
	\centering
	\caption{Comprehensive comparison with state-of-the-art methods on both image quality and latency.
		Image quality is evaluated using five metrics: PSNR, SSIM, APD, LPIPS, and FID.
		The \textit{Time} is obtained via the average watermarking time over 1,000 samples.
		All experiments are conducted on either an Intel Xeon CPU or an NVIDIA RTX 3090 GPU.
		More comparisons are shown in Tab. \ref{tab:time_detail}.)}
	\label{tab:psnr}
	\resizebox{0.98\linewidth}{!}{%
		\begin{tabular}{@{}ccccccccccccc@{}}
			\toprule
			\multirow{2}{*}{Method} & \multicolumn{5}{c}{COCO} & \multicolumn{5}{c}{ImageNet} & \multicolumn{2}{c}{Time ($\times 10^{-3}$ s) $\downarrow$} \\ \cmidrule(lr){2-6} \cmidrule(lr){7-11} \cmidrule(lr){12-13}
			& PSNR (dB) $\uparrow$ & SSIM $\uparrow$ & APD $\downarrow$ & LPIPS $\downarrow$ & FID $\downarrow$ & PSNR (dB) $\uparrow$ & SSIM $\uparrow$ & APD $\downarrow$ & LPIPS $\downarrow$ & FID $\downarrow$ & CPU & GPU \\ \midrule
			FIN \cite{flow-based} & 30.93 & 0.769 & 5.185 & 0.209 & 145.45 & 30.68 & 0.782 & 5.316 & 0.196 & 129.57 & 82.43 & 20.23 \\
			DWSF \cite{dwsf}& 28.01 & 0.908 & 5.332 & 0.096 & 25.54 & 27.95 & 0.912 & 5.260 & 0.091 & 20.90 & 146.1 & 33.91 \\
			MuST \cite{must}& 25.60 & 0.724 & 9.840 & 0.236 & 119.17 & 25.58 & 0.730 & 9.845 & 0.233 & 114.68 & 18.58 & 6.57 \\
			UDH \cite{udh}& 34.26 & 0.914 & 4.899 & 0.083 & 22.24 & 34.25 & 0.911 & 4.899 & 0.112 & 21.23 & 1.05 & 1.52 \\
			\textbf{HiWL} (Ours) & \textbf{37.86} & \textbf{0.969} & \textbf{1.822} & \textbf{0.048} & \textbf{17.06} & \textbf{37.84} & \textbf{0.970} & \textbf{1.827} & \textbf{0.050} & \textbf{16.47} & \textbf{1.03} & \textbf{1.41} \\ \bottomrule
		\end{tabular}%
	}
\end{table*}

\textbf{Baselines.}
We conduct comparative analyses with FIN \cite{flow-based}, DWSF \cite{dwsf}, UDH \cite{udh}, and MuST \cite{must}. It is worth noting that both UDH and our method belong to the single-shot paradigm for watermark generation. In contrast, DWSF, FIN, and MuST are latent-based methods.

\begin{figure*}[t]
	\centering
	\includegraphics[width=\linewidth]{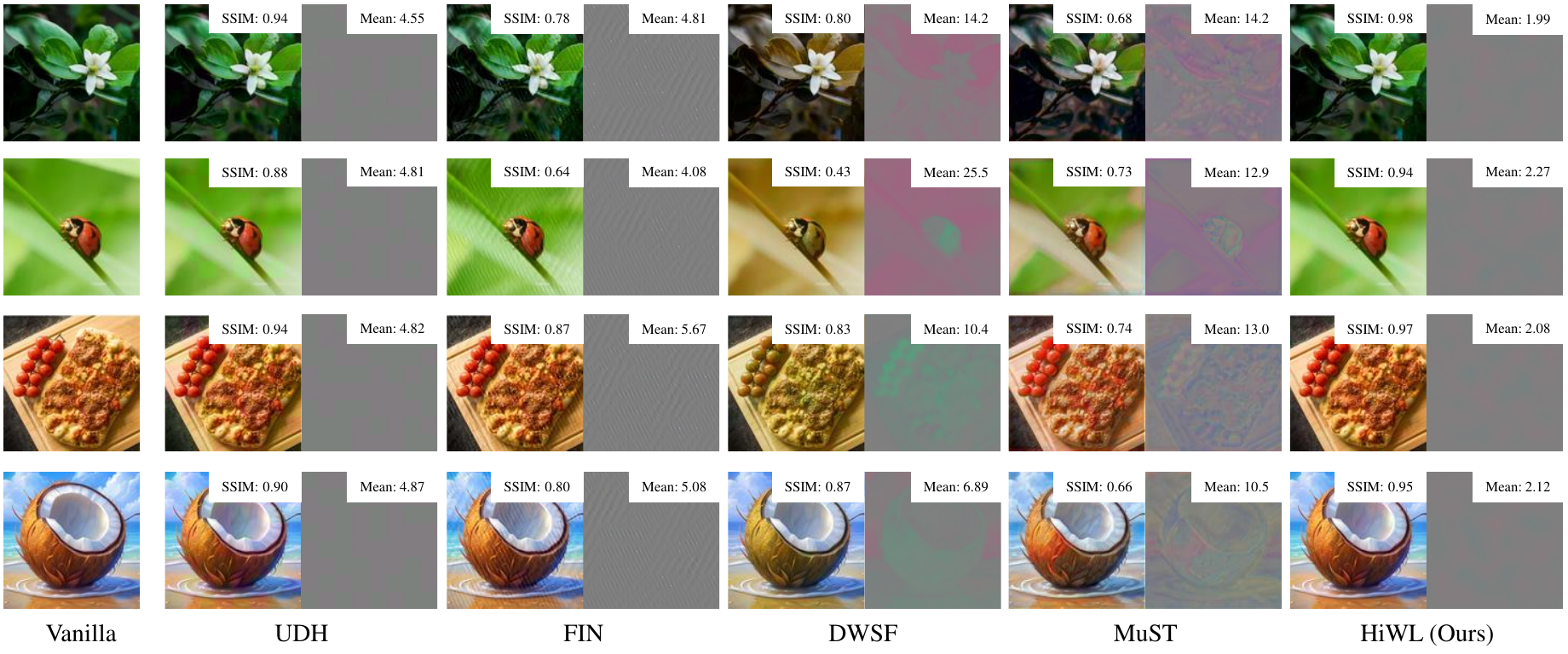}
	\caption{Visualization of watermarked images. Each method, except for \textit{vanilla}, is paired with the watermark images (left) and the residual images (right). The SSIM metric is tested between the \textit{vanilla} and the corresponding watermark images.} 
\label{fig:visualization}
\end{figure*}

\textbf{Implementation Details and Metrics.}
We implement our method using PyTorch and train it on 2 $\times$ NVIDIA RTX A6000 GPU.
We use AdamW \cite{adamw} as the optimizer and set the learning rate to 1e-4. 
We train for a total of 200 epochs, with 100 epochs in both the first and second stages. The image size used for both training and testing across all methods is 3$\times$128$\times$128, and the message length is 64 bits.
In addition, we adopt a wide range of distortions to simulate potential real-world scenarios. These comprise 18 different types of noise distortions, including JPEG compression, Gaussian noise (GN), Gaussian filter (GF), dropout, median blur, color variation, brightness variation, saturation variation, hue variation, contrast variation, resize, cropping, picture-in-picture (PIP), padding, occlusion, rotation, shear, and affine transformations. The effects of these distortions are shown in Fig.~\ref{fig:noise}. The training and testing parameters are listed in Tab.~\ref{tab:para_noise}.

To evaluate the robustness of a watermark model, we use Bit Accuracy (\%) to assess the similarity between the pre-defined message $M$ and the extracted message $M^{'}$. The accuracy is obtained by calculating the Bit Error Rate (BER):
\begin{equation}
\mathrm{Accuracy}
= 1 - \mathrm{BER}
= \left( 1 - \frac{1}{n} \sum_{i=1}^{n} \left( M_i \oplus M_i' \right) \right) \times 100\% ,
\label{eq:acc}
\end{equation}
where $\oplus$ is the exclusive or operation between bits and $n$ is the number of bits in the message. 
To investigate the invisibility, Peak Signal-to-Noise Ratio (PSNR), Structural Similarity Index (SSIM), Average Pixel Difference (APD), Fréchet Inception Distance (FID) \cite{fid}, and Learned Perceptual Image Patch Similarity (LPIPS) \cite{lpips} are employed to measure the invisibility of the generated watermark.
Among these metrics, PSNR, SSIM, Wass, and APD are traditional evaluation metrics, while FID and LPIPS are widely adopted deep feature-based metrics for assessing image quality. 
For broad applicability, the time behind watermark generation is leveraged.

\begin{table*}[t]
\centering
\caption{Investigation under 18 noise distortions. Four Datasets (DS) refer to DRCT \cite{drct}, GenImage \cite{genimage}, ImageNet \cite{imagenet}, and HumanArt \cite{humanart}.}
\label{tab:robustness}
\resizebox{0.98\linewidth}{!}{%
	\begin{tabular}{@{}ccccccccccc@{}}
		\toprule
		Method & JPEG & GN & GF & Dropout & MF & Color & Bright & Contrast & Hue & Saturation \\ \midrule
		FIN \cite{flow-based}& \textbf{99.94\%} & \textbf{99.99\%} & 99.32\% & 99.87\% & 53.54\% & 99.00\% & 98.26\% & \textbf{94.83\%} & \textbf{99.99\%} & \textbf{99.98\%} \\
		DWSF \cite{dwsf}& 97.03\% & 99.97\% & \textbf{99.98\%} & 99.27\% & 95.29\% & 97.63\% & 96.80\% & 92.05\% & 85.68\% & 99.23\% \\
		MuST \cite{must}& 73.85\% & 77.99\% & 73.87\% & 76.67\% & 65.23\% & 75.50\% & 75.57\% & 73.40\% & 71.42\% & 72.91\% \\
		UDH \cite{udh}& 98.32\% & 99.02\% & 98.19\% & 97.52\% & 96.11\% & 93.90\% & 88.69\% & 81.62\% & 96.22\% & 85.42\% \\
		\textbf{HiWL} (Ours) & 98.46\% & 99.93\% & 99.83\% & \textbf{99.94\%} & \textbf{98.08\%} & \textbf{99.76\%} & \textbf{98.38\%} & 94.53\% & 99.94\% & 99.02\% \\ \midrule
		& Resize & Crop & PIP & Padding & Occlusion & Rotate & Shear & Affine & \textbf{AVG} & \textbf{AVG (4 DS)} \\ \midrule
		FIN \cite{flow-based}& \textbf{99.90\%} & 55.03\% & 50.04\% & 53.40\% & 94.19\% & 50.01\% & 65.97\% & 50.31\% & 81.31\% & 81.29\% \\
		DWSF \cite{dwsf}& 99.99\% & 71.22\% & 72.71\% & 96.46\% & 98.38\% & 67.14\% & 99.11\% & 64.28\% & 90.68\% & 91.17\% \\
		MuST \cite{must}& 77.29\% & 69.60\% & 68.05\% & 75.21\% & 77.67\% & 72.07\% & 77.35\% & 75.50\% & 73.84\% & 73.26\% \\
		UDH \cite{udh}& 98.94\% & 52.97\% & 50.09\% & 53.77\% & 89.98\% & 49.66\% & 59.22\% & 50.98\% & 80.03\% & 80.43\% \\
		\textbf{HiWL} (Ours) & 99.71\% & \textbf{93.10\%} & \textbf{92.99\%} & \textbf{97.99\%} & \textbf{99.87\%} & \textbf{99.83\%} & \textbf{99.33\%} & \textbf{98.90\%} & \textbf{98.31\%} & \textbf{98.00\%} \\ \bottomrule
	\end{tabular}%
}
\end{table*}

\subsection{Main Results}
\begin{figure*}[t]
\centering
\includegraphics[width=0.96\linewidth]{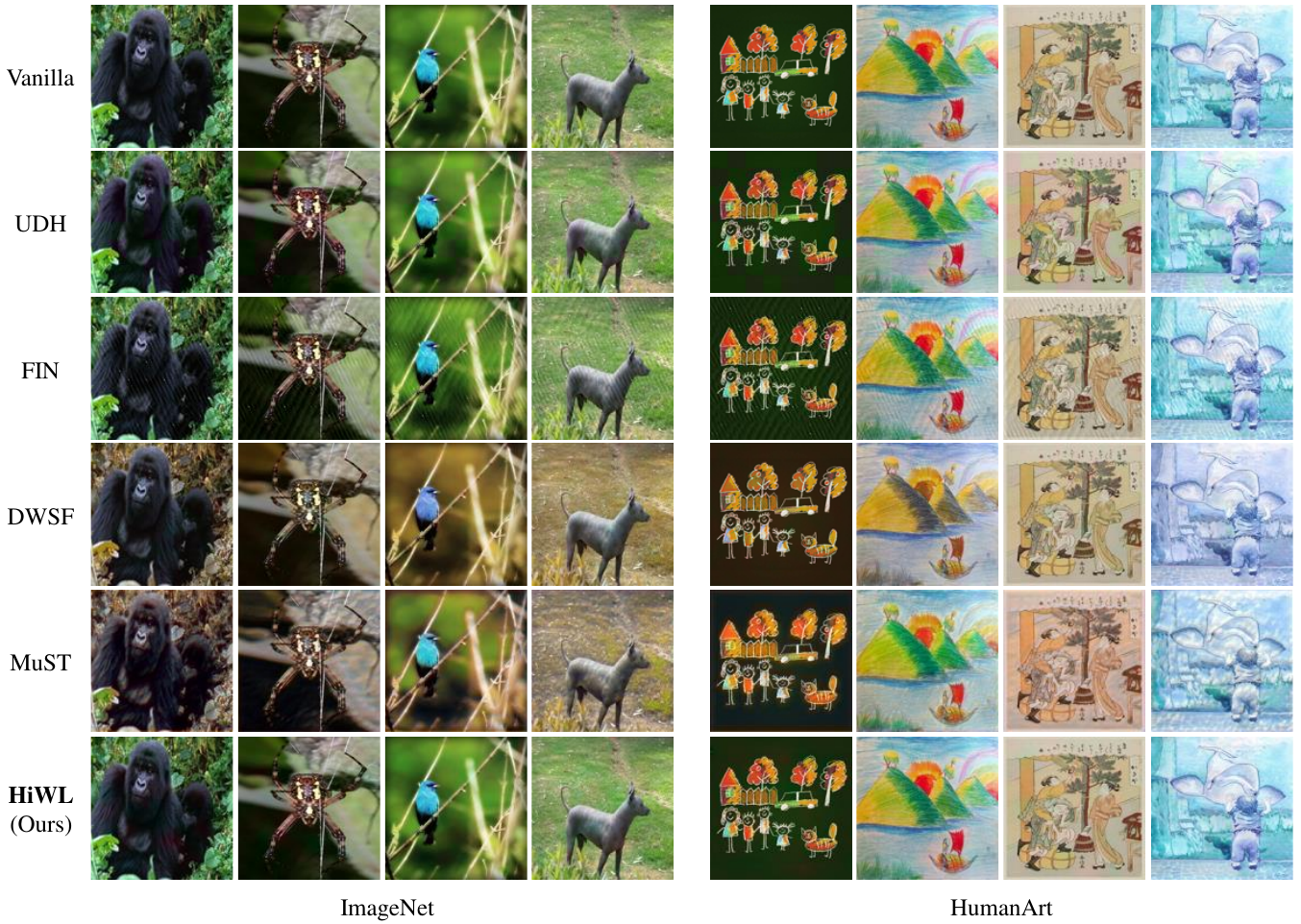}
\caption{Visualization of watermarked images from ImageNet \cite{imagenet} and HumanArt \cite{humanart}.}
\label{fig:visualization-1}
\end{figure*}

\begin{figure*}[b]
\centering
\includegraphics[width=0.96\linewidth]{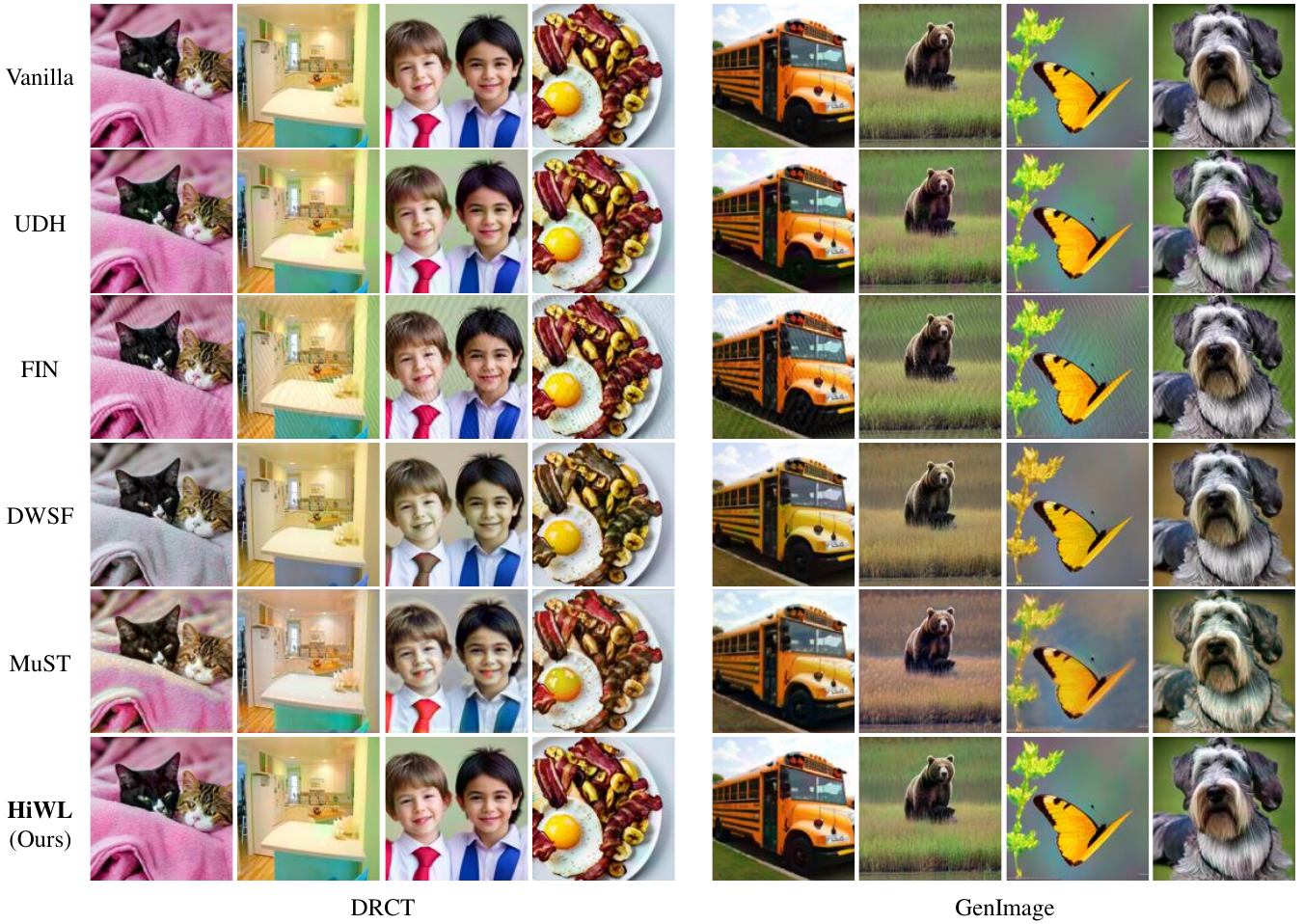}
\caption{Visualization of watermarked images from GenImage \cite{genimage} and DRCT \cite{drct}.}
\label{fig:visualization-2}
\end{figure*}

\textbf{Investigation of Invisibility.}
The qualitative visualization of watermark invisibility is shown in Fig.~\ref{fig:visualization}, and the quantitative evaluation results are presented in the medium column of Tab.~\ref{tab:psnr}. We also visualize watermarked images using four different datasets, as shown in Fig.~\ref{fig:visualization-1} and Fig.~\ref{fig:visualization-2}. From the results, we find that HiWL achieves the best performance across all metrics, surpassing the second-best (UDH) by a significant margin. In terms of PSNR and SSIM, it achieves \textbf{37.86} dB and \textbf{0.969}, respectively, while UDH records 34.26 dB and 0.914. 
Additionally, in terms of APD, HiWL achieves \textbf{1.822}, which is less than half of UDH’s 4.899. When performing evaluation on ImageNet, DRCT, and GemImage datasets, HiWL still maintains similar improvements. 
This indicates that our method exhibits excellent imperceptibility.

\begin{table}[t]
\centering
\caption{Investigation of watermarking efficiency for different watermarking methods. All tests are performed on either an Intel Xeon CPU or an NVIDIA RTX 3090 GPU, with results averaged over 1K executions.
	\textit{Load} time denotes loading images from disk into memory; \textit{Save} time denotes writing images from memory to disk; and Peripheral Component Interconnect Express (\textit{PCIe}) time denotes data transfer between host and GPU memory.
}
\label{tab:time_detail}
\resizebox{0.92\textwidth}{!}{%
	\begin{tabular}{ccccccc}
		\toprule
		Model & FLOPs & Load ($\times 10^{-5}$ s) & PCIe ($\times 10^{-5}$ s) & Watermark ($\times 10^{-5}$   s) & Save ($\times 10^{-5}$ s) & Device \\ \midrule
		\multirow{2}{*}{FIN \cite{flow-based}} & \multirow{2}{*}{1187.42M} & 41.10 & \textbackslash{} & 8139.63 & 62.38 & CPU \\ \cmidrule(l){3-7} 
		&  & 69.10 & 32.42 & 1857.42 & 63.97 & GPU \\ \midrule
		\multirow{2}{*}{MuST \cite{must}} & \multirow{2}{*}{5262.14M} & 41.36 & \textbackslash{} & 1746.94 & 69.49 & CPU \\ \cmidrule(l){3-7} 
		&  & 44.31 & 40.64 & 499.03 & 73.18 & GPU \\ \midrule
		\multirow{2}{*}{DWSF \cite{dwsf}} & \multirow{2}{*}{24348.77M} & 56.91 & \textbackslash{} & 14490.39 & 65.29 & CPU \\ \cmidrule(l){3-7} 
		&  & 59.97 & 12.28 & 3249.99 & 68.97 & GPU \\ \midrule
		\multirow{2}{*}{UDH \cite{udh}} & \multirow{2}{*}{49152} & 58.24 & \textbackslash{} & 3.24 & 43.26 & CPU \\ \cmidrule(l){3-7} 
		&  & 67.28 & 33.20 & 3.02 & 48.20 & GPU \\ \midrule
		\multirow{2}{*}{\textbf{HiWL}} & \multirow{2}{*}{\textbf{49152}} & 57.90 & \textbackslash{} & 3.45 & 41.20 & CPU \\ \cmidrule(l){3-7} 
		&  & 63.85 & 33.20 & 2.86 & 41.02 & GPU \\ \bottomrule
	\end{tabular}%
}
\end{table}

\textbf{Investigation of Broad Applicability.}
In the single-shot setting, an RGB watermark is pre-generated, and watermarking time is measured by embedding it into 1,000 images. In contrast, latent-based methods require the model to be called 1,000 times to watermark those images, as shown on the right side of Tab.~\ref{tab:psnr}. 
We further report detailed watermarking time breakdowns, including FLOPs, image loading
time, and data transfer time between host memory and GPU memory, as shown in Tab.~\ref{tab:time_detail}.
We use the fvcore library \cite{fvcore} to measure FLOPs for all methods except UDH and HiWL, since it automatically ignores minor addition operations. For UDH and HiWL, the FLOPs are manually computed based on the addition of images.
The results indicate that the watermarking latency of single-shot methods is
lower than the image loading latency. Consequently, HiWL achieves performance
comparable to UDH while significantly outperforming latent-based methods,
demonstrating strong applicability across diverse scenarios.

\textbf{Investigation of Robustness.}
We test all methods using their respective paradigms, where HiWL and UDH are evaluated under the single-shot paradigm, while the others follow latent-based paradigms. As shown in Tab.~\ref{tab:robustness}, HiWL consistently outperforms UDH under all of the 18 noise attacks, achieving an average performance of \textbf{98.3\%}, while UDH records 80.0\%. The best-performing latent-based method, DWSF, achieves an average accuracy of only 90.7\%. Among the 18 noise attacks, HiWL achieves the best performance in 11 cases. 
This indicates that our method's robustness remains guaranteed, even after undergoing Stage 2 of generalization training.

\begin{figure*}[t]
\centering
\includegraphics[width=\linewidth]{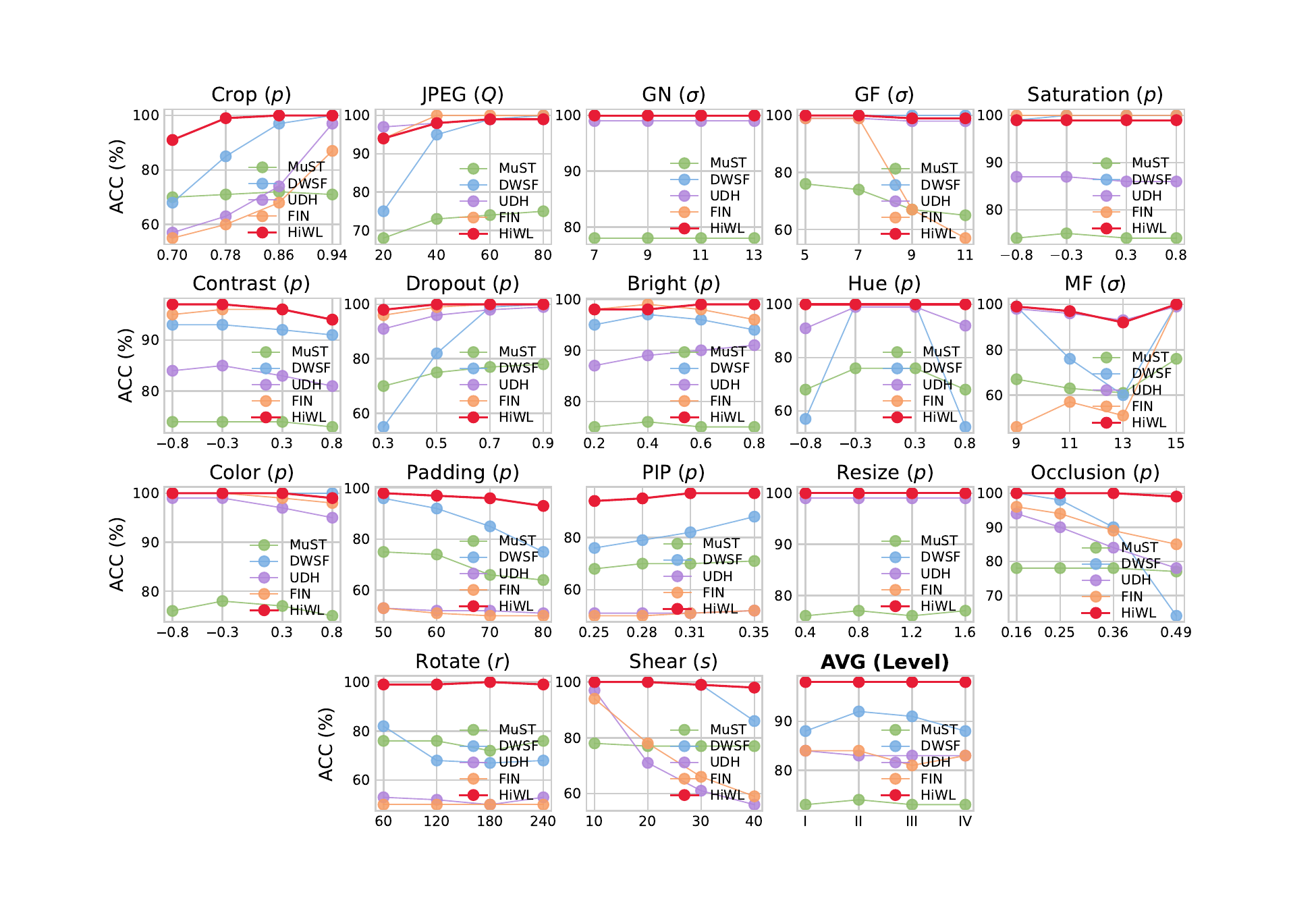}
\caption{Comparison of Bit Accuracy (\%) under 18 noise factors with four
different levels. AVG is the average value of these noises corresponding to different
levels.}
\label{fig:noise-factor}
\end{figure*}

We also evaluate the robustness under varying noise factors, as shown in Fig.~\ref{fig:noise-factor}. HiWL still achieves stable advantages under different levels of noise.

To further evaluate robustness from a broader perspective, we test DWSF, FIN, and MuST under a single-shot setting. As shown in Tab. \ref{tab:srobustness}, HiWL significantly outperforms the second-best method, FIN, which achieves 79.3\%.

\textbf{Robustness under Residual Attack.}
Generalizable watermarks are not directly coupled with image features. This will result in similar structures across watermark images. One of the most challenging attacks for the generalized watermark is the RGB residual removal attack. Attackers can exploit this by removing potential RGB residuals to launch the attack. 
Specifically, for each test batch, we simulate an attacker's acquisition of an RGB watermark by first generating an RGB residual from a cover image and its corresponding watermark image. This generated residual then serves as the attack watermark, which is subsequently subtracted from a separate watermarked image. The results are shown in Tab.~\ref{tab:residualattack}.
UDH processes only the messages without incorporating the image latent space, resulting in an average accuracy of 66.9\% under the attack. 
In comparison, HiWL can embed watermarks more covertly in the general latent space, leading to better performance under attack, achieving an average accuracy of 78.6\%.

\begin{figure}
\centering
\includegraphics[width=0.65\linewidth]{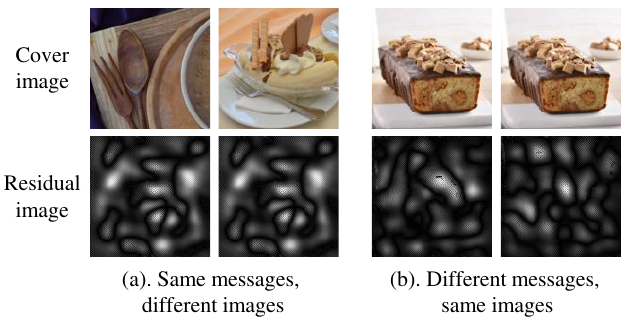}
\caption{Visualization of normalized RGB watermarks.}
\label{fig:shape}
\end{figure}

\subsection{Shape of Generalized RGB watermark}
We randomly select two cover images and embed them with the same message. Evaluating the resulting watermark residuals reveals identical watermark shapes, as shown in Fig.~\ref{fig:shape}(a). Conversely, when we apply the same image with different watermark residuals, the results differ, as depicted in Fig.~\ref{fig:shape}(b). This demonstrates that the RGB watermark residual is message-specific and fully validates the effectiveness of our method.

\subsection{Robustness under Domain Transfer}
In this section, we investigate the watermark model under a cross-domain setting, where we apply an RGB watermark generated on images from one dataset to images from a distinctly different dataset. 
As shown in Tab.~\ref{tab:domain} and Tab.~\ref{tab:0-255}, HiWL is unaffected across different domains in the process of generating RGB watermarks. However, applying these generated RGB watermarks to different domains does introduce a certain degree of impact. Specifically, directly adding the watermark to the new cover image, which operates on a 0/255 scale, leads to an overflow issue. This degradation primarily occurs with noise that significantly increases brightness or contrast. This is because perturbations close to the threshold, like a value of 252, become saturated to 255 after undergoing such brightness noise, consequently causing the watermark to fail. Our method, however, still demonstrates strong robustness across domains.

\begin{table}[htbp]
\centering
\caption{Investigation under cross-domain transfer. Line 1: Generate an RGB watermark using a normal dataset (COCO) and add it to images from a new domain. Line 2: Generate an RGB watermark in a different domain and apply it to COCO images.}
\label{tab:0-255}
\resizebox{0.55\linewidth}{!}{
	\begin{tabular}{cccc}
		\toprule
		& sketch \cite{humanart} & cartoon \cite{humanart}& drct \cite{drct}\\ \midrule
		$I_c \to domain$ & 95.48\% & 97.80\% & 98.54\% \\ 
		$domain \to I_c^{'}$ & 98.67\% & 98.55\% & 98.59\% \\
		\bottomrule
	\end{tabular}
}
\end{table}

\begin{table}[htbp]
\centering
\caption{Comparison among state-of-the-art methods under different domains.}
\label{tab:domain}
\resizebox{0.4\linewidth}{!}{
	\begin{tabular}{ccc}
		\toprule
		Method & 0 & 255 \\ \midrule
		FIN \cite{flow-based}& 79.54\% & 77.51\% \\ 
		DWSF \cite{dwsf}& 85.09\% & 84.48\% \\
		MuST \cite{must}& 52.20\% & 52.55\% \\
		UDH \cite{udh}& 79.41\% & 76.05\% \\
		$I_c \to domain$ (Ours)& 89.49\% & 88.17\% \\
		$domain \to I_c^{'}$ (Ours)& 98.67\% & 98.62\% \\
		\bottomrule
	\end{tabular}
}
\end{table}

\begin{table}[t]
\centering
\caption{Reporting Bit Accuracy (\%) under 18 noise distortions with RGB residual attack. All methods adopt a single-shot paradigm. }
\label{tab:residualattack}
\resizebox{0.7\linewidth}{!}{%
	\begin{tabular}{@{}ccc|ccc@{}}
		\toprule
		Noise & UDH & \textbf{HiWL} (Ours) & Noise & UDH & \textbf{HiWL} (Ours) \\ \midrule
		JPEG & 75.63\% & \textbf{76.23\%} & Contrast & 69.91\% & \textbf{82.43\%} \\
		GN & 77.09\% & \textbf{81.64\%} & Resize & 76.54\% & \textbf{77.98\%} \\
		GF & 75.67\% & \textbf{78.66\%} & Crop & 52.52\% & \textbf{67.72\%} \\
		Dropout & 76.74\% & \textbf{85.90\%} & PIP & 49.76\% & \textbf{74.15\%} \\
		MF & \textbf{75.93\%} & 74.29\% & Padding & 51.88\% & \textbf{78.44\%} \\
		Color & 75.89\% & \textbf{90.51\%} & Occlusion & 72.09\% & \textbf{87.95\%} \\
		Bright & 73.86\% & \textbf{87.69\%} & Rotate & 49.34\% & \textbf{61.61\%} \\
		Saturation & 71.48\% & \textbf{86.61\%} & Shear & 55.02\% & \textbf{67.64\%} \\
		Hue & 75.63\% & \textbf{87.34\%} & Affine & 49.05\% & \textbf{67.90\%} \\ \bottomrule
	\end{tabular}%
}
\end{table}

\begin{table}[t]
\centering
\caption{Investigating the invisibility and robustness at different training stages. PSNR and APD represent invisibility, while Dropout and JPEG represent robustness. \textbf{L} indicates that the latent-based paradigm is used to test, and \textbf{S} indicates that the single-shot paradigm is used to test.}
\label{tab:abtion}
\resizebox{0.82\linewidth}{!}{%
\begin{tabular}{ccc|cc|ccc}
\toprule
\multicolumn{2}{c}{Train} & Test & \multirow{2}{*}{PSNR (dB)} & \multirow{2}{*}{APD} & \multirow{2}{*}{Dropout} & \multirow{2}{*}{JPEG} & AVG \\
Stage 1 & Stage 2 & L/S &  &  &  &  & (18 distortions) \\ \midrule
\ding{51} &  & L & 34.70 & 2.636 & 99.87\% & 98.82\% & 98.86\% \\
\ding{51} &  & S & 34.86 & 2.586 & 82.60\% & 77.20\% & 79.02\% \\
& \ding{51} & L & 35.51 & 0.274 & 63.91\% & 63.93\% & 61.17\% \\
& \ding{51} & S & 35.57 & 0.273 & 63.79\% & 63.85\% & 61.10\% \\ 
\ding{51} & \ding{51} & L & 37.84 & 1.826 & 99.96\% & 98.82\% & 98.30\% \\
\ding{51} & \ding{51} & S (Ours) & 37.86 & 1.822 & 99.94\% & 98.46\% & 98.31\% \\ \bottomrule
\end{tabular}%
}
\end{table}

\begin{table*}[tbp]
\centering
\caption{Comparison of Bit Accuracy (\%) under 18 noise distortions. All methods use the single-shot paradigm to generate watermarks. $*$ denotes latent-based watermarking methods while employing a single-shot paradigm for generalized watermark generation. \textbf{4 Dataset (DS)} refers to DRCT \cite{drct}, GenImage \cite{genimage}, ImageNet \cite{imagenet}, and HumanArt \cite{humanart}.}
\label{tab:srobustness}
\resizebox{0.98\linewidth}{!}{%
\begin{tabular}{@{}ccccccccccc@{}}
\toprule
Method & JPEG & GN & GF & Dropout & MF & Contrast & Bright & Color & Hue & Saturation \\ \midrule
FIN $^{*}$ \cite{flow-based}& 98.10\% & 99.11\% & 97.66\% & 98.04\% & 54.84\% & 90.48\% & 95.90\% & 97.62\% & 99.09\% & \textbf{99.14\%} \\
DWSF $^{*}$ \cite{dwsf}& 51.75\% & 52.71\% & 51.95\% & 52.29\% & 50.80\% & 60.48\% & 52.79\% & 52.93\% & 51.90\% & 60.11\% \\
MuST $^{*}$ \cite{must}& 62.64\% & 71.40\% & 62.46\% & 68.60\% & 55.11\% & 64.81\% & 66.89\% & 67.57\% & 65.66\% & 64.04\% \\
UDH \cite{udh}& 98.32\% & 99.02\% & 98.19\% & 97.52\% & 96.11\% & 81.62\% & 88.69\% & 93.90\% & 96.22\% & 85.42\% \\
\textbf{HiWL} (Ours) & \textbf{98.46\%} & \textbf{99.93\%} & \textbf{99.83\%} & \textbf{99.94\%} & \textbf{98.08\%} & \textbf{94.53\%} & \textbf{98.38\%} & \textbf{99.76\%} & \textbf{99.94\%} & 99.02\% \\ \midrule
& Resize & Crop & PIP & Padding & Occlusion & Rotate & Shear & Affine & \textbf{AVG} & \textbf{AVG (4 DS)} \\ \midrule
FIN $^{*}$ \cite{flow-based}& 98.75\% & 53.99\% & 50.09\% & 53.40\% & 92.86\% & 50.23\% & 65.14\% & 50.48\% & 79.25\% & 80.51\% \\
DWSF $^{*}$ \cite{dwsf}& 52.11\% & 50.12\% & 52.28\% & 58.20\% & 52.02\% & 49.92\% & 52.13\% & 50.33\% & 53.05\% & 52.73\% \\
MuST $^{*}$ \cite{must}& 69.77\% & 60.61\% & 58.40\% & 65.66\% & 71.25\% & 60.68\% & 64.89\% & 65.02\% & 64.58\% & 64.57\% \\
UDH \cite{udh}& 98.94\% & 52.97\% & 50.09\% & 53.77\% & 89.98\% & 49.66\% & 59.22\% & 50.98\% & 79.22\% & 80.43\% \\
\textbf{HiWL} (Ours) & \textbf{99.71\%} & \textbf{93.10\%} & \textbf{92.99\%} & \textbf{97.99\%} & \textbf{99.87\%} & \textbf{99.83\%} & \textbf{99.33\%} & \textbf{98.90\%} & \textbf{98.22\%} & \textbf{98.00\%} \\ \bottomrule
\end{tabular}%
}
\end{table*}

\subsection{Ablation Study}
In this section, we thoroughly investigate the necessity of two-stage training. Specifically, we conduct an experimental analysis by testing the robustness of models focusing on three different training strategies: Stage 1 only, Stage 2 only, and the combined Stage 1 + Stage 2. All three branches are trained for the same number of epochs. The distribution alignment corresponds to Stage 1, while the generalized representation learning corresponds to Stage 2. All results are shown in Tab.~\ref{tab:abtion}. We observe that training with Stage 1 only achieves good robustness (L test), but its generalization ability is weaker (S test). When a strong penalty with wide multi-image adaptation is applied across cover images in Stage 2, we achieve a significant improvement in generalization while maintaining the same robustness and higher invisibility as Stage 1.

It's worth noting that even without generalization training (Stage 2), the model doesn't completely fail in S tests (line 2, Tab.~\ref{tab:abtion}). This strongly supports our view that image and message features are only weakly coupled, as there's no loss function explicitly forcing their fusion. Therefore, even results trained under a latent-based approach still exhibit a degree of generalization.

When comparing Stage 2 with the combination of Stage 1 and Stage 2, we find that the combined training significantly outperforms Stage 2 alone in both latent-based (L) and single-shot (S) evaluations. This suggests that training Stage 2 in isolation is insufficient. From a modeling perspective, we believe that Stage 2 learns to generalize the representations established in Stage 1; therefore, training a generalized representation directly is ineffective.


\FloatBarrier  

\section{Conclusion}
In this work, we present HiWL, a novel Hierarchical Watermark Learning framework designed to address the fundamental limitations of existing learning-based watermarking methods. By decomposing the training process into two complementary stages, HiWL effectively balances the key requirements of practical watermarking systems.
Specifically, in the first stage, we leverage distribution alignment to establish a weak yet meaningful coupling between the watermark and the host image in the latent space. This design allows the learned watermark to achieve strong robustness while remaining imperceptible, ensuring reliable message recovery without compromising visual quality.
Building upon this foundation, the second stage introduces a multi-image adaptation strategy with asynchronous optimization, which significantly enhances the generalization capability of the watermarking model. As a result, HiWL exhibits broad applicability across diverse images and transformation scenarios, overcoming the limited adaptability commonly observed in prior methods.
Overall, HiWL demonstrates that robustness, invisibility, and broad applicability can be jointly achieved within a unified learning framework. We believe this work offers valuable insights into generalized watermarking, an essential yet challenging research direction, and may inspire future efforts toward more flexible and scalable watermarking systems.

\section*{Acknowledgments}
This research do not receive any specific grant from funding agencies in the public, commercial, or not-for-profit sectors.

\bibliographystyle{elsarticle-num} 
\bibliography{references} 

\end{document}